\documentclass[a4paper,fleqn]{cas-dc}

\usepackage[numbers]{natbib}

\def\tsc#1{\csdef{#1}{\textsc{\lowercase{#1}}\xspace}}
\tsc{WGM}
\tsc{QE}
\tsc{EP}
\tsc{PMS}
\tsc{BEC}
\tsc{DE}

\usepackage{placeins}

\usepackage{amsmath,amssymb,amsfonts}
\usepackage{algorithmic}
\usepackage{graphicx}
\usepackage{algorithm,algorithmic}
\usepackage{hyperref}
\hypersetup{hidelinks=true}
\usepackage{textcomp}

\usepackage{bm}

\usepackage{pifont}
\usepackage{nicefrac}
\usepackage{siunitx}

\usepackage[dvipsnames]{xcolor}
\usepackage{wrapfig}
\usepackage{tikz}
\usepackage{adjustbox}
\usetikzlibrary{calc}
\usepackage{import}
\usetikzlibrary{quotes,arrows.meta}
\usetikzlibrary{positioning}

\usepackage{Box}
\usepackage{Box_inside}
\usepackage{Box_inside_stacked}

\definecolor{newgreen}{rgb}{0.67, 0.82, 0.57}
\definecolor{newred}{rgb}{0.9, 0.09, 0.1}
\definecolor{newblue}{rgb}{0.56, 0.67, 0.85}
\usepackage{booktabs}
\usepackage{arydshln}
\usepackage{multirow}
\usepackage{hyperref}
\hypersetup{hidelinks}
\makeatletter
\def\adl@drawiv#1#2#3{%
        \hskip.5\tabcolsep
        \xleaders#3{#2.5\@tempdimb #1{1}#2.5\@tempdimb}%
                #2\z@ plus1fil minus1fil\relax
        \hskip.5\tabcolsep}
\newcommand{\cdashlinelr}[1]{%
  \noalign{\vskip\aboverulesep
           \global\let\@dashdrawstore\adl@draw
           \global\let\adl@draw\adl@drawiv}
  \cdashline{#1}
  \noalign{\global\let\adl@draw\@dashdrawstore
           \vskip\belowrulesep}}
\newcommand{\cdashlinelrdotted}[1]{%
  \noalign{\vskip\aboverulesep
           \global\let\@dashdrawstore\adl@draw
           \global\let\adl@draw\adl@drawiv}
  \cdashline{#1}[.4pt/1pt]
  \noalign{\global\let\adl@draw\@dashdrawstore
           \vskip\belowrulesep}}
\makeatother

\let\printorcid\relax 

\begin{document}
\let\WriteBookmarks\relax
\def\floatpagepagefraction{1}
\def\textpagefraction{.001}
\shorttitle{MSA-CNN: A Lightweight Multi-Scale CNN with
Attention}
\shortauthors{S. Goerttler et~al.}

\title [mode = title]{MSA-CNN: A Lightweight Multi-Scale CNN with Attention for Sleep Stage Classification}                      
\tnotemark[1]

\tnotetext[1]{This manuscript has been accepted for publication in Biomedical Signal Processing and Control. 
This version reflects changes made during peer review and is not the final published version. 
The final version of record is available at: \href{https://doi.org/10.1016/j.bspc.2026.110141}{https://doi.org/10.1016/j.bspc.2026.110141}}


\author[1,2]{Stephan Goerttler}[style=english]

\credit{Conceptualization of this study, Methodology, Software, Validation,Writing - Original Draft, Visualization}

\affiliation[1]{organization={Centre for Computational Science and Mathematical Modelling, Coventry University},
                addressline={Priory St}, 
                city={Coventry},
               citysep={}, 
                postcode={CV1 5FB}, 
                country={United Kingdom}}
\affiliation[2]{organization={Institute for Infocomm Research, A*STAR},
                addressline={1 Fusionopolis Way \#21-01 Connexis}, 
                city={},
               citysep={}, 
                postcode={138632}, 
                country={Singapore}}
\author[2]{Yucheng Wang}[style=english]
\credit{Software, Writing - Review \& Editing}
\author[2]{Emadeldeen Eldele}[%
   ]
\credit{Conceptualization, Writing - Review \& Editing}
\author[1]{Fei He}[style=english]
\credit{Writing - Review \& Editing, Supervision, Funding acquisition}
\cormark[1]
\ead{fei.he@coventry.ac.uk}
\author[2]{Min Wu}[style=english]
\credit{Writing - Review \& Editing, Supervision, Funding acquisition}
\cormark[1]
\ead{wumin@a-star.edu.sg}






\cortext[cor1]{Corresponding authors.}


\begin{abstract}
Recent advancements in machine learning-based signal analysis, coupled with open data initiatives, have fuelled efforts in automatic sleep stage classification. 
Despite the proliferation of classification models, few have prioritised reducing model complexity, which is a crucial factor for practical applications.
In this work, we introduce Multi-Scale and Attention Convolutional Neural Network (MSA-CNN), a lightweight architecture featuring as few as $\sim$10,000 parameters. 
MSA-CNN leverages a novel multi-scale module employing complementary pooling to eliminate redundant filter parameters and dense convolutions. Model complexity is further reduced by separating temporal and spatial feature extraction and using cost-effective global spatial convolutions.
This separation of tasks not only reduces model complexity but also mirrors the approach used by human experts in sleep stage scoring.
We evaluated both small and large configurations of MSA-CNN against nine state-of-the-art baseline models across three public datasets, treating univariate and multivariate models separately. 
Our evaluation, based on repeated cross-validation and re-evaluation of all baseline models, demonstrated that the large MSA-CNN outperformed all baseline models on all three datasets in terms of accuracy and Cohen's kappa, despite its significantly reduced parameter count.
Lastly, we explored various model variants and conducted an in-depth analysis of the key modules and techniques, providing deeper insights into the underlying mechanisms.
The code for our models, baselines, and evaluation procedures is available at \url{https://github.com/sgoerttler/MSA-CNN}.
\end{abstract}

\begin{keywords}
Convolutional neural networks \sep electroencephalography \sep multivariate signals \sep sleep stage classification
\end{keywords}

\maketitle

\section{Introduction}
\label{sec:introduction}
\begin{figure*}[t!]
    \centering
    \adjustbox{trim=0 17pt 0 0, clip}{%
    \resizebox{0.95\textwidth}{!}{%
    \input{graphics/networks_general.tikz}%
    }%
    }
    \caption{Full architecture of our proposed Multi-Scale and Attention Convolutional Neural Network (MSA-CNN). The Multi-Scale Module (MSM, see Figure \ref{fig:multiscale}) extracts high-level spectro-morphological features from the input sleep epoch. Subsequently, a global spatial convolution detects co-activation patterns across all input channels, yielding time-dependent feature tokens. These tokens are then passed to our Temporal Context Module (TCM, see Figure \ref{fig:attention}), which adjusts the meaning of each token depending on the surrounding context. The time average of these tokens is then interpreted by means of a fully connected layer, yielding the final classification of the input signal.}
    \label{fig:model}
\end{figure*}
Sleep is an important function of human physiology, essential for overall health and well-being. Sleep disorders, which encompass a wide range of conditions that affect sleep quality, timing, and duration, can significantly impact daily functioning and quality of life.
Today, polysomnography is used for assessing sleep and sleep disorders, allowing for comprehensive monitoring of various physiological parameters during sleep \cite{rundo2019polysomnography}.
In polysomnography, the classification of the recordings into five sleep stages serves as an important tool to understand the sleep architecture.
The gold standard in sleep stage classification involves manual assessment by sleep experts, who rely on visual inspection of recorded data \cite{sekkal2022automatic}.
However, fuelled by advances in machine learning (ML) and the growing number of publicly available datasets, computer-based sleep stage classification has made significant strides in recent years \cite{phan2022automatic}.

Many ML-based sleep stage classification models are designed for single-channel EEG inputs, owing to their simplicity in application and independence from electrode montage configurations \cite{supratak2017deepsleepnet,perslev2021u,zhou2022singlechannelnet}.
In contrast, emerging studies increasingly focus on exploiting the rich spatial information in multi-channel recordings, employing intricate methods such as graph convolutional networks (GCNs) \cite{jia2021multi,ji2022jumping,wang2024fully,ma2025mfsleepnet}, transformer encoders \cite{dai2023multichannelsleepnet,zhou2025enhancing}, and 3D convolutional
neural networks (CNNs) \cite{ji20233dsleepnet}.
However, the resulting models often contain large numbers of parameters, making them susceptible to overfitting and limiting practical applicability \cite{hoge2018primer}.
While numerous studies have explored strategies to reduce the complexity of univariate models \cite{fiorillo2021deepsleepnet,zhou2021lightsleepnet,jadhav2022automated,yang2023lwsleepnet,cheng2023eeg,zhou2024interpretable}, including implementations on hardware-constrained devices \cite{yao2023cnn,liu2023micro}, relatively few have focused on the complexity of multivariate models \cite{chambon2018deep,lawhern2018eegnet,lin2023multimodal}.
Model complexity is also closely tied to explainability, which is a
critical factor for the clinical deployment of sleep stage classification models. 
Recent studies have employed gradient-weighted class activation mapping to explain weight-based attention \cite{liu2023micro} or to approximate input importance \cite{vaquerizo2023explainable,dutt2023sleepxai}; yet, to the best of our knowledge, explanations for transformer-based attention remain absent. 

The primary aim of this work is to develop a lightweight and interpretable Multi-Scale and Attention CNN (MSA-CNN) that effectively incorporates spatial information (see Section \ref{sec:proposed_model}).
To achieve this, we employ three complexity reduction strategies in parallel.
Firstly, we introduce a novel \textit{complementary pooling} technique, which enables multi-scale feature learning using a compact filter for each scale (see Subsection \ref{ssec:msm}). This technique reduces model complexity by eliminating redundant filter parameters and forms the core of our proposed two-layer Multi-Scale Module (MSM).
In contrast, previous multi-scale model designs for sleep stage classification \cite{jia2021multi,supratak2017deepsleepnet} and EEG-based brain-computer interface \cite{ko2021multi,roy2022efficient} 
typically increase the filter size\thinspace---\thinspace sometimes to several hundred weights\thinspace---\thinspace or use atrous convolutions to expand the filter's receptive field. 
Our second technique, based on Chambon et al. \cite{chambon2018deep}, involves separating temporal and spatial feature extraction, which reduces complexity in a similar way to depthwise separable convolutions \cite{kaiser2017depthwise}.
This separation also mimics manual sleep stage classification, where sleep experts detect temporal patterns for each channel before interpreting the co-activation of these patterns across the channels \cite{iber2007aasm}.
As our final complexity reduction technique, we employ 
a global spatial convolutional layer to capture spatial dependencies. This approach is particularly cost-effective for the small number of channels typical in sleep stage classification. 
Conceptually, global spatial convolutions are equivalent to fully connected layers across both spatial and feature dimensions, enabling holistic processing of graph-structured data.
Importantly, they are simpler and more computationally efficient than common spatial feature extraction methods,
such as graph convolutional layers \cite{jia2020graphsleepnet,jia2021multi,ji2022jumping} or graph pooling layers \cite{wang2023multivariate}. 
Apart from these complexity reduction techniques, our model also features an attention-based Temporal Context Module (TCM) to capture long-range dependencies, which builds on previous research demonstrating the potential of attention mechanisms for sleep stage classification \cite{eldele2021attention,shen2023lgsleepnet,pan2024causalattennet}.
In Subsection \ref{ssec:TCM}, we introduce a visualisation tool that illustrates incoming and outgoing attention of the TCM, further enhancing the interpretability of our model.

Our experimental setup is described in Section \ref{sec:exp_setup}. We developed a small MSA-CNN configuration with $\sim$10,000 parameters and a large configuration with $\sim$40,000 parameters, designed for use with larger datasets.
Both models were tested on three public datasets, and their performance was compared against nine state-of-the-art (SOTA) baseline models, categorised as either univariate or multivariate. 
To foreshadow our main results in Section \ref{sec:comparison}, we found that the large lightweight MSA-CNN significantly outperformed all tested SOTA models, despite its substantial reduction in model complexity.
We further evaluate the efficacy of the modules and techniques using ablation studies and parameter sensitivity analyses. 
Regarding model explainability, we provide an example that demonstrates how self-attention modifies the interpretation of a sleep spindle in the presence of a K-complex during non-REM sleep stage N2. 
Section \ref{sec:summary_conclusion} gives a summary and conclusion of this study.

The key contributions of this work are as follows:
\begin{itemize}
    \item We propose and validate a novel multi-scale temporal feature extraction module based on complementary pooling. The module reduces model complexity by eliminating redundant parameters.
    \item We design a multivariate, lightweight sleep stage classification architecture (MSA-CNN) with as few as $\sim$10,000 parameters using the proposed MSM module and the attention-based TCM module. We validate the multivariate approach and provide insights into the attention mechanism.
    \item We test several MSA-CNN configurations and re-evaluate nine sleep stage classification baselines on three datasets using a comprehensive validation strategy. The results demonstrate the superiority of our model, both in terms of performance and model complexity.
\end{itemize}

\section{Proposed Model}
\label{sec:proposed_model}

\subsection{Overall Architecture}
Figure \ref{fig:model} illustrates our proposed MSA-CNN, which is inspired by the way sleep experts classify sleep epochs \cite{iber2007aasm}.
First, an offset and a scale are learnt for each channel of the input sleep epoch, serving as a trainable preprocessing step.
The proposed MSM module then extracts spectro-morphological features, such as K-complexes, sleep spindles, or alpha waves, across a broad range of frequency scales. Subsequently, a global spatial convolutional layer captures the spatial dependencies, with the spatial filters spanning all channels.
The result of this convolution is a time-resolved feature map that encodes spatial co-activation of characteristic brainwave patterns. By viewing the feature vector at each time-step as a token, we can apply transformer-like attention using the TCM. This module detects contextual information in the signal by adjusting the interpretation of each feature vector based on its surrounding context. These feature vectors are then averaged across time and processed by a fully connected layer. Finally, a softmax operation generates the class probabilities, serving as the final output of the model.

\subsection{Multi-Scale Module (MSM)}
\label{ssec:msm}
The MSM, shown in Figure \ref{fig:multiscale}, employs a simple, yet powerful technique to regularise filter-based feature extraction on multiple scales. 
The core idea is to split a temporal convolution into multiple convolutions, each tasked with extracting features on a different time scale.
To this end, the convolutions are encased by an input pooling and a complementary pooling operation for each scale. 
The preceding pooling along time determines the scale of the convolutional filter by defining its receptive field and resolution in the input signal.
This operation is then followed by a temporal convolution, which extracts low-level spectro-morphological features on the respective scale. Crucially, the filter size of this convolution can be kept small for each scale. Note also that the filters can be either unimodal, where they are shared across all channels, or multimodal, where each filter is applied separately to individual channels. To maintain the shape of the pooled input, the padding type is set to ``same''. For sufficiently long input signals, the resulting border effects can be neglected.
Finally, a complementary pooling operation downsamples the feature maps of the convolution to a common scale, allowing the downsampled feature maps to be merged. 
Specifically, the feature map size can be matched by defining the complementary pooling factor as $p_\mathrm{comp} = p_\mathrm{tot} / p_\mathrm{in}$, where $p_\mathrm{in}$ is the input pooling and $p_\mathrm{tot}$ is the combined total pooling. 

A second temporal convolution integrates the scales to form high-level spectro-morphological features. 
It is plausible that these features encode the characteristic brainwave patterns used by sleep experts to interpret the physiological data.

\begin{figure*}[tbh]
    \centering
    \resizebox{0.75\textwidth}{!}{%
    \input{graphics/networks_CP.tikz}%
    }
    \caption{Illustration of our Multi-Scale Module (MSM) with four scales. Each scale pathway consists of a pooling operation and a convolutional layer, followed by a complementary pooling operation to match the feature map shape across scales.
    The receptive field of each convolution is shaded in the input signal. 
    A second temporal convolution integrates all four scales.
    }
    \label{fig:multiscale}
\end{figure*}
\begin{figure}[]
    \centering
    \resizebox{\columnwidth}{!}{%
    \input{graphics/networks_attention.tikz}%
    }
    \caption{Temporal context module using multi-head self-attention. Multi-head attention (Equation \ref{eq:mha}) requires the computation of query (blue), key (green), and value (red) maps. The sequence of attention and feed-forward layer (grey area) is carried out $N_{lay}$ times.
    %
    }
    \label{fig:attention}
\end{figure}

\subsection{Temporal Context Module (TCM)}
\label{ssec:TCM}
\color{black}We leverage multi-head self-attention \cite{vaswani2017attention} to learn contextual information in the data, based on the work of Eldele et al. \cite{eldele2021attention}. The TCM is shown in Figure \ref{fig:attention}. Firstly, the feature vectors at each time step are embedded into a lower-dimensional space of dimension $d_{emb}$ via a linear layer, yielding a more compact feature representation. Secondly, positional information is embedded in each feature at time $t$ and dimension $i$ by adding the following positional encoding:
\begin{align}
\text{PE}(t, i) =
    \begin{cases}
    \sin\left(\frac{t}{10,000^{\nicefrac{i}{d_{emb}}}}\right) & \text{if } i \text{ is even}, \\
    \cos\left(\frac{t}{10,000^{\nicefrac{(i-1)}{d_{emb}}}}\right) & \text{if } i \text{ is odd}.
    \end{cases}
\end{align}

The embedded features are then transformed into query ($Q$), key ($K$), and value ($V$) representations through linear projections. The number of representations for each category is given by the number of heads $N_{hd}$, while the transformed representations have dimension $d_k=d_{emb}/N_{hd}$.
The attention weights for each head $h$ are given by:
\begin{align}
    \label{eq:attention}
    A_h = \mathrm{softmax}\left(\frac{Q_hK_h^\top}{\sqrt{d_k}}\right).
\end{align}
The attention weights, together with the value matrices $V_h$, are then used to compute the multi-head attention:
\begin{align}
    \label{eq:mha}
    A_{mh} = \text{Concat}\left(..., A_h V_h, ...\right) W^O,
\end{align}
where $W^O$ is a learnt square weight matrix.
This attention mechanism enables the embedded features to adapt their meaning depending on the context. A residual connection added to the output allows the input to bypass the attention mechanism, which helps to preserve information and facilitate training. A layer norm operation is then applied to the output.

Subsequently, the features pass through a feed-forward network given by two fully connected layers, each followed by a rectified linear unit (ReLU) activation function and dropout. The dimension of the first layer is twice that of the input dimension, while the second layer reduces the representation back down to the input dimension. A residual connection is incorporated into the output of the feed-forward network.
The sequence of multi-head attention and feed-forward network is repeated $N_{lay}$ times.

We further propose a visualisation tool to illustrate the incoming and outgoing attention employed by the TCM for a given sleep epoch. First, the participant associated with this sample is excluded from the training set. The sample is then processed through the trained network, and the attention weight matrix $A_h$, defined in Equation \ref{eq:attention}, is retrieved from the first multi-head attention operation.
This weight matrix, specific to the sample, enables the calculation of the mean \textit{incoming attention} as the average across the first dimension. This definition of incoming attention quantifies the extent to which each point in time is attended to by all other points. 
Conversely, \textit{outgoing attention}, defined here at the point of maximum incoming attention, is obtained by slicing $A_h$ along its first dimension at this point. It therefore represents the degree to which each point attends to the most attended point.

\FloatBarrier

\section{Experimental Setup}
\label{sec:exp_setup}
\subsection{Datasets}

\setlength{\tabcolsep}{3pt}
\begin{table}[]
    \centering
    \caption{Key figures of datasets used in experiment}
    \resizebox{1\columnwidth}{!}{
    \begin{tabular}{lcccccccc}
\toprule
dataset & channels & subjects &  \multicolumn{5}{c}{class samples} &  total\\\cmidrule(lr){4-8}
 &\smash{\raisebox{0.5em}{(used)}} &&W&N1&N2&N3&REM & \smash{\raisebox{0.5em}{samples}}\\ 
\midrule
\multirow{2}*{ISRUC-S3} & \multirow{2}*{10 (9)} & \multirow{2}*{10} & 1,674& 1,217& 2,616& 2,016& 1,066&\multirow{2}*{8,589} \\ 
& & & \textit{19.5\%}& \textit{14.2\%}& \textit{30.5\%}& \textit{23.5\%}& \textit{12.4\%}& \\\addlinespace[3pt]
\multirow{2}*{Sleep-EDF-20} & \multirow{2}*{6 (4)} & \multirow{2}*{20} & 8,285&  2,804& 17,799&  5,703&  7,717&\multirow{2}*{42,308}\\ 
 & & & \textit{19.6\%}& \textit{6.6\%}& \textit{42.1\%}& \textit{13.5\%}& \textit{18.2\%}& \\\addlinespace[3pt]
\multirow{2}*{Sleep-EDF-78} & \multirow{2}*{6 (4)} & \multirow{2}*{78} & 65,951& 21,522& 69,132& 13,039& 25,835&\multirow{2}*{195,479}\\ 
 & & & \textit{33.7\%}& \textit{11.0\%}& \textit{35.4\%}& \textit{6.7\%}& \textit{13.2\%}& \\
\bottomrule
\end{tabular}}
    \label{tab:dataset}
\end{table}
\setlength{\tabcolsep}{6pt}
The publicly available datasets used in this study are ISRUC-S3, Sleep-EDF-20, and Sleep-EDF-78. A summary of the three datasets is given in Table \ref{tab:dataset}.
The first dataset, \textbf{ISRUC-S3}, was recorded by Khalighi et al. from 10 healthy subjects during sleep \cite{khalighi2016isruc}. The recordings were divided into 30-second epochs, resulting in a total of 8,589 samples. These samples were classified by two human experts into five sleep stages according to the AASM standard \cite{iber2007aasm}. The dataset, downsampled to 100\,Hz, comprises preprocessed physiological recordings from six referenced EEG channels, two electrooculography (EOG) channels, one electromyogram (EMG), and one electrocardiogram (ECG). We discarded the ECG from our proposed MSA-CNN model due to its periodic nature, which sets it apart from the other channels. 
To eliminate high-frequency noise, such as power line interference, we preprocessed each channel by applying a fourth-order low-pass Butterworth filter with a cutoff frequency of 40\,Hz.

The second and third labelled polysomnographic datasets, \textbf{Sleep-EDF-20} and \textbf{Sleep-EDF-78}, are sourced from PhysioBank \cite{goldberger2000physiobank}. These datasets were collected from healthy participants aged 25 to 101 over two nights as part of a study investigating the effects of age on sleep. The Sleep-EDF-20 dataset comprises 20 subjects and has overall 42,308 labelled 30-second samples after removing samples with artefacts. The extended Sleep-EDF-78 dataset includes 78 subjects and has as many as 195,479 labelled samples with artefacts removed. 
All samples were manually scored according to Rechtschaffen \& Kales (R\&K) using the two EEG channels \cite{kales1968manual}.
The recorded signals consist of two referenced EEG channels (Fpz-Cz and Pz-Oz), one EOG channel, and one EMG. EEG and EOG were sampled at 100\,Hz, while EMG was recorded at 1\,Hz and subsequently upsampled to match the other modalities. Two additional physiological signals\thinspace---\thinspace air-nasal flow and rectal temperature\thinspace---\thinspace were excluded from this study and all baseline experiments due to their slowly varying nature.
Similarly to the ISRUC-S3 dataset, we preprocessed the data with a 40\,Hz low-pass filter.

\subsection{Baseline Models}
\label{ssec:validation}
We evaluate our proposed model against nine SOTA ML-based sleep stage classification models on the ISRUC-S3, Sleep-EDF-20, and Sleep-EDF-78 datasets. 
Our model selection is based on availability of code, relevance to our model, novelty, and impact in the field.
The selected models include two univariate models, five multivariate models, and three models that can be either univariate or multivariate.
The nine baseline models can be briefly characterised as follows:
\begin{itemize}
    \item \textbf{DeepSleepNet} \cite{supratak2017deepsleepnet} is an early univariate model that combines a CNN with a bidirectional-long short-term memory neural network. 
    \item \textbf{EEGNet} \cite{lawhern2018eegnet} is a CNN for EEG classification which incorporates depthwise and separable convolutions to reduce the amount of model parameters. It can be used with either univariate or multivariate inputs.
    \item \textbf{AttnSleep} \cite{eldele2021attention} is a univariate model that features a multi-resolution CNN and a temporal context encoder based on multi-head attention.
    \item \textbf{HierCorrPool} \cite{wang2023multivariate} is a univariate or multivariate graph neural network designed to capture hierarchical channel correlations.
    \item \textbf{FC-STGNN} \cite{wang2024fully} is a graph neural network which models the spatial-temporal dependencies between the channels. It can be used with univariate or multivariate inputs.
    \item \textbf{GraphSleepNet} \cite{jia2020graphsleepnet} is a multivariate GCN that uses only engineered features as input.
    \item \textbf{MSTGCN} \cite{jia2021multi} is a multivariate model consisting of a multi-resolution CNN-based feature extractor followed by a separately trained GCN.
    \item \textbf{JK-STGCN} \cite{ji2022jumping} is a multivariate model that builds on the previous MSTGCN and similarly comprises a CNN-based feature extractor and a GCN. 
    \item \textbf{cVAN} \cite{yang2024cvan}, or cross-view alignment network, is a recent multivariate residual-like neural network with scale-aware attention.
\end{itemize}

\subsection{Model Settings and Variants}
\label{ssec:modelconfig}
In this study, we developed a small and a large MSA-CNN, which can be univariate, multivariate, or multivariate with multimodal filters. An overview of the set of parameters for all model variants can be found in Table \ref{tab:hyperparameters}.
Table \ref{tab:scales} shows the settings and key figures for the MSM scales used in our experiment. 

To begin with, the number of MSM scales was set to 4 to adequately cover the relevant spectral range in EEG. We use average pooling for the input pooling operation, while using max pooling for the complementary pooling operation to accentuate filter activations.
The multi-scale convolution (MSM I) has a temporal kernel size of 15, while the scale-integration convolution (MSM II) has a kernel size of 5.
In the multivariate configuration, the spatial convolution has a kernel which spans the entire spatial dimension.

The small MSA-CNN is extended to the large model by doubling the number of filters in the MSM II layer (from 16 to 32 in the multivariate configuration) and in the spatial layer (from 32 to 64).
In the TCM, the embedding dimension is increased from 16 to 32, the number of heads is doubled from 2 to 4, and the number of layers is doubled from 1 to 2.
Note that the number of filters in the lower-level multi-scale convolution remains the same for the small and the large MSA-CNN, with a value of 8 for the unimodal variants.

\setlength{\tabcolsep}{4.5pt}
\begin{table}[]
    \centering
    \caption{Model parameters for MSA-CNN small and MSA-CNN large, configured as univariate, multivariate, or multimodal. Kernel sizes are specified by their spatial and temporal dimensions}
    \resizebox{1\columnwidth}{!}{
    \begin{tabular}{llcccccc}
\toprule
layer & hyperparameter & \multicolumn{3}{c}{MSA-CNN (small)}& \multicolumn{3}{c}{MSA-CNN (large)}\\ \cmidrule(lr){3-5}\cmidrule(lr){6-8}
 &  & univar. & multivar. & multimod. & univar. & multivar. & multimod.\\
\midrule
MSM I & \# scales & &\multicolumn{1}{c}{--\enspace4\enspace--}& & &\multicolumn{1}{c}{--\enspace4\enspace--}&\\
 & kernel size & &\multicolumn{1}{c}{--\enspace1$\times$15\enspace--}& & &\multicolumn{1}{c}{--\enspace1$\times$15\enspace--}&\\
& \# filters / scale & 8&8&4*$N_{ch}$& 8&8& 4*$ N_{ch}$\\
MSM II & kernel size & &\multicolumn{1}{c}{--\enspace1$\times$5\enspace--}&& &\multicolumn{1}{c}{--\enspace1$\times$5\enspace--}& \\
 & \# filters & 16&16&8*$N_{ch}$ & 32&32&16*$N_{ch}$\\ 
 spatial & kernel size & 1$\times$5 &$N_{ch}\times 1$&$N_{ch}\times 1$ & 1$\times$5 &$N_{ch}\times 1$&$N_{ch}\times 1$\\
& \# filters & &\multicolumn{1}{c}{--\enspace32\enspace--}& & &\multicolumn{1}{c}{--\enspace64\enspace--}&\\ 
TCM & embedding dim.& &\multicolumn{1}{c}{--\enspace16\enspace--}& & &\multicolumn{1}{c}{--\enspace32\enspace--}&\\
 & \# heads& &\multicolumn{1}{c}{--\enspace2\enspace--}& & &\multicolumn{1}{c}{--\enspace4\enspace--}&\\
 & \# layers & &\multicolumn{1}{c}{--\enspace1\enspace--}& & &\multicolumn{1}{c}{--\enspace2\enspace--}&\\
\bottomrule
\end{tabular}
}
    \label{tab:hyperparameters}
\end{table}
\begin{table}[]
    \centering
    \caption{Scale settings and characteristics in the experiment}
    \resizebox{1\columnwidth}{!}{
    \begin{tabular}{lcccccc}
\toprule
scale &colour- & pooling& compl.&receptive &frequency & frequency \\
& code & size&pooling& field [ms] &range [Hz] & spacing [Hz]\\
\midrule
scale I & blue & 1 & 8 &150&[0, 46.7]&6.7 \\
scale II & green & 2 & 4 & 300&[0, 23.3] & 3.3 \\
scale III & yellow & 4 & 2 & 600 & [0, 11.7] & 1.7 \\
scale IV & gray & 8 & 1 & 1,200 & [0, \phantom{0}5.8] & 0.8 \\
\bottomrule
\end{tabular}}
    \label{tab:scales}
\end{table}

The univariate model variant is derived from the multivariate variant by replacing the spatial convolution with a temporal convolution with filter size 5. On the other hand, the multimodal variant is implemented by applying filters for each channel separately. To balance the additional number of parameters with the extraction capability, we set the number of filters per scale and per channel to 4 in layer MSM I, and to 8 and 16 in layer MSM II for the small and large models, respectively.

\subsection{Model training}
\label{ssec:training}
All MSA-CNN variants were trained for 100 epochs using the Adam optimiser \cite{KingBa15}.
The learning rate for the small models was set to 0.001, while the learning rate for the more complex large models was reduced to 0.0001 to facilitate training. In the case of the multivariate and multimodal variants,
we increased the learning rate for the higher-level TCM and output layer to 0.001 to accelerate convergence, as preliminary experiments indicated that a higher rate led to faster and more stable convergence.
Light regularisation was applied using a dropout rate of 0.1 and weight decay of 0.0001.
The baseline models were trained using the hyperparameters and configurations specified in the respective publications.

\subsection{Validation}
We validated our model using 10-fold repeated subject-wise cross-validation, which is more robust than traditional validation strategies used in automatic sleep stage classification.
The importance of repetitions is underscored by the limited number of subjects, which reduces the effective size of polysomnography datasets substantially.
We set the number of repetitions for the ISRUC-S3 and the Sleep-EDF-20 datasets to 10, and for the larger Sleep-EDF-78 dataset to 3 due to computational constraints.
Although it is common practice to shuffle folds during cross-validation repetitions, the limited number of subjects leads to significant overlap between folds across shuffled iterations. We therefore fix the folds across repetitions to ensure consistency.

We re-evaluated all baselines using our comprehensive validation strategy, ensuring a fair comparison between the models. 
To prevent data leakage, we strictly separated the training and test sets during hyperparameter optimisation, where applicable. Specifically, we modified the early stopping criteria in the baselines \cite{jia2020graphsleepnet,jia2021multi,ji2022jumping}, and \cite{yang2024cvan} to monitor the training set rather than the test set.
We slightly deviated from our proposed validation procedure in the case of the HierCorrPool and FC-STGNN models, where we reduced the number of repetitions for the ISRUC-S3 and the Sleep-EDF-20 datasets from 10 to 3 due to computational constraints.
For transparency and replicability, we provide the code and validation procedures both for our model as well as all baseline models.\footnote{\url{https://github.com/sgoerttler/MSA-CNN}}

\subsection{Evaluation Metrics}
We evaluate model performance using three metrics, namely accuracy, macro F1 score, and Cohen's kappa. Importantly, we compute the metric for each fold separately before taking the average weighted by the number of test fold samples. The final metric is given by the average across all repetitions. 

Our first evaluation metric is accuracy, which is a straightforward and widely adopted measure that reflects the overall proportion of correct predictions. The accuracy over $K$ classes on a given test fold is computed as:
\begin{align}
    \text{Accuracy} = \frac{\sum_{i=1}^{K} TP_i + TN_i}{N_s},
\end{align}
where $TP_i$ and $TN_i$ are the respective number of true positives and true negatives for class $i$, while $N_s$ denotes the total number of samples.

We secondly report the macro F1 score, which we include due to the significant class imbalance in the datasets, especially in the Sleep-EDF datasets. The macro F1 score assigns equal weight to all classes, ensuring that minority classes are not overshadowed by the majority. 
It can be given in terms of the precision and recall as follows:
\begin{align}
    \label{eq:MF1}
    MF1 &= \frac{1}{K}\sum_{i=1}^{K} 2 \cdot \frac{\text{Precision}_i \cdot \text{Recall}_i}{\text{Precision}_i + \text{Recall}_i},\\
    \text{Precision}_i &= \frac{TP_i}{TP_i + FP_i},\quad
    \text{Recall}_i = \frac{TP_i}{TP_i + FN_i}.
\end{align}
Here, $FP_i$ and $FN_i$ denote the respective number of false positives and false negatives of class $i$. Note that the sum in equation \ref{eq:MF1} does not contain variable weights, thereby giving an equal weight of $1/K$ to all classes.

Lastly, we include \textbf{Cohen's kappa} as our third evaluation metric \cite{cohen1960coefficient}. This metric accounts for chance performance, which is particularly relevant in the presence of highly imbalanced data. 
A value of 1 signifies perfect agreement between the manually rated labels and the model classifier, while a value of 0 signifies chance agreement.

To evaluate model complexity, we firstly consider the number of trainable parameters. In addition, we report the \textbf{MFLOPS}, which me measure using the ptflops package \cite{ptflops} for models implemented in PyTorch and the TensorFlow Profiler for models implemented in Keras.
Note that two models with the same number of parameters can have different MFLOPS model complexity, for example due to variations in the computational cost of convolutional operations induced by different pooling configurations.

\begin{table*}[]
    \centering
    \caption{Comparison of model performance against state-of-the-art models. The performance is presented as the average across all repetitions, with the variability indicated by the standard deviation. Bold indicates the highest evaluation metric for each mode, dataset, and metric, while underlined indicates the second highest metric}
    \resizebox{0.9\textwidth}{!}{
    \begin{tabular}{lll@{\hspace{12pt}}c@{\hspace{6pt}}c@{\hspace{6pt}}c@{\hspace{12pt}}c@{\hspace{6pt}}c@{\hspace{6pt}}c@{\hspace{12pt}}c@{\hspace{6pt}}c@{\hspace{6pt}}c}
\toprule
mode & model & year & \multicolumn{3}{c}{\hspace{-12pt}ISRUC-S3} & \multicolumn{3}{c}{\hspace{-12pt}Sleep-EDF-20} & \multicolumn{3}{c}{\hspace{-12pt}Sleep-EDF-78} \\ \cmidrule(lr{18pt}){4-6}\cmidrule(lr{18pt}){7-9}\cmidrule(lr{9pt}){10-12}
& & & accuracy & macro F1 & kappa& accuracy & macro F1 & kappa& accuracy & macro F1 & kappa \\
\midrule
\multirow{7}*{\rotatebox[origin=c]{90}{univariate}} & DeepSleepNet \cite{supratak2017deepsleepnet}& 2017 & 73.3$\,\pm\,$0.6&68.1$\,\pm\,$0.6&64.7$\,\pm\,$0.8 & 80.7$\,\pm\,$0.2&71.9$\,\pm\,$0.3&73.4$\,\pm\,$0.2 & 77.0$\,\pm\,$0.2&67.3$\,\pm\,$0.2&67.3$\,\pm\,$0.3\\
& EEGNet \cite{lawhern2018eegnet}& 2018 & 71.2$\,\pm\,$0.7&65.0$\,\pm\,$0.8&62.2$\,\pm\,$0.9 & 79.5$\,\pm\,$0.5&66.1$\,\pm\,$0.7&71.8$\,\pm\,$0.6 & 76.0$\,\pm\,$0.4&62.8$\,\pm\,$0.3&65.5$\,\pm\,$0.6 \\
& AttnSleep \cite{eldele2021attention}& 2021 & 73.4$\,\pm\,$0.8&\underline{69.0$\,\pm\,$0.7}&65.0$\,\pm\,$1.0 & 81.3$\,\pm\,$0.4&\textbf{74.9}$\,\bm{\pm}\,$\textbf{0.3}&\underline{74.4$\,\pm\,$0.4} & \underline{79.9$\,\pm\,$0.2}&\textbf{71.0}$\,\bm{\pm}\,$\textbf{0.1}&\underline{71.6$\,\pm\,$0.2} \\
& HierCorrPool \cite{wang2023multivariate}& 2023 & 68.0$\,\pm\,$2.1&63.2$\,\pm\,$2.2&58.6$\,\pm\,$2.2 & 77.3$\,\pm\,$1.9&68.9$\,\pm\,$1.7&68.7$\,\pm\,$2.5 & 75.3$\,\pm\,$0.4&66.2$\,\pm\,$0.4&65.3$\,\pm\,$0.5 \\
& FC-STGNN \cite{wang2024fully}& 2024 & 64.6$\,\pm\,$2.0&58.6$\,\pm\,$1.8&53.4$\,\pm\,$2.2 & 73.9$\,\pm\,$0.0&65.6$\,\pm\,$0.3&63.9$\,\pm\,$0.2 & 66$\,\pm\,$11&54$\,\pm\,$13&52$\,\pm\,$15 \\
& \textbf{MSA-CNN (small)} & 2024 & \textbf{75.2}$\,\bm{\pm}\,$\textbf{0.6}&\textbf{70.5}$\,\bm{\pm}\,$\textbf{0.6}&\textbf{67.2}$\,\bm{\pm}\,$\textbf{0.7} & \underline{81.5$\,\pm\,$0.4}&72.2$\,\pm\,$0.6&74.3$\,\pm\,$0.6 & 79.2$\,\pm\,$0.3&68.0$\,\pm\,$0.5&70.1$\,\pm\,$0.4 \\
& \textbf{MSA-CNN (large)} & 2024 & \underline{73.7$\,\pm\,$0.6}&67.8$\,\pm\,$0.7&\underline{65.2$\,\pm\,$0.8} & \textbf{82.2}$\,\bm{\pm}\,$\textbf{0.3}&\underline{72.9$\,\pm\,$0.5}&\textbf{75.2}$\,\bm{\pm}\,$\textbf{0.4} & \textbf{80.1}$\,\bm{\pm}\,$\textbf{0.0}&\textbf{71.0}$\,\bm{\pm}\,$\textbf{0.2}&\textbf{71.7}$\,\bm{\pm}\,$\textbf{0.0} \\
\midrule
\multirow{9}*{\rotatebox[origin=c]{90}{multivariate}} & EEGNet \cite{lawhern2018eegnet}& 2018 & 75.9$\,\pm\,$1.0&71.9$\,\pm\,$1.0&68.2$\,\pm\,$1.3 & 82.7$\,\pm\,$0.3&74.9$\,\pm\,$0.6&76.1$\,\pm\,$0.4 & 78.0$\,\pm\,$0.8&67.0$\,\pm\,$1.1&68.6$\,\pm\,$1.1 \\
& GraphSleepNet \cite{jia2020graphsleepnet}& 2020 & 69.8$\,\pm\,$1.7&66.7$\,\pm\,$2.1&60.3$\,\pm\,$2.1 & 77.4$\,\pm\,$0.7&66.1$\,\pm\,$1.1&68.5$\,\pm\,$1.0 & 69.2$\,\pm\,$2.6&55.5$\,\pm\,$3.4&56.2$\,\pm\,$4.1 \\
& MSTGCN \cite{jia2021multi}& 2021 & 77.2$\,\pm\,$0.6&73.5$\,\pm\,$0.7&69.7$\,\pm\,$0.7 & 82.5$\,\pm\,$0.3&75.9$\,\pm\,$0.4&75.8$\,\pm\,$0.4 & 80.7$\,\pm\,$0.2&\textbf{74.7}$\,\bm{\pm}\,$\textbf{0.2}&73.3$\,\pm\,$0.3 \\
& JK-STGCN \cite{ji2022jumping}& 2022 & 75.9$\,\pm\,$0.6&72.0$\,\pm\,$0.4&68.1$\,\pm\,$0.7 & 83.3$\,\pm\,$0.4&76.6$\,\pm\,$0.4&76.8$\,\pm\,$0.6 & \underline{81.0$\,\pm\,$0.1}&\underline{73.9$\,\pm\,$0.8}&\underline{73.4$\,\pm\,$0.3} \\
& HierCorrPool \cite{wang2023multivariate}& 2023 & 72.1$\,\pm\,$1.3&67.8$\,\pm\,$1.6&63.1$\,\pm\,$1.7 & 81.1$\,\pm\,$0.4&74.8$\,\pm\,$0.8&73.9$\,\pm\,$0.6 & 80.6$\,\pm\,$0.3&73.1$\,\pm\,$0.2&72.7$\,\pm\,$0.3 \\
& FC-STGNN \cite{wang2024fully}& 2024 & 68.6$\,\pm\,$0.4&64.6$\,\pm\,$1.0&58.5$\,\pm\,$0.8 & 77.8$\,\pm\,$0.3&70.5$\,\pm\,$0.3&69.4$\,\pm\,$0.4 & 76.2$\,\pm\,$0.1&68.1$\,\pm\,$0.3&66.8$\,\pm\,$0.2 \\
& cVAN \cite{yang2024cvan}& 2024 & 72.5$\,\pm\,$2.0&67.7$\,\pm\,$2.5&63.7$\,\pm\,$2.5 & 81.6$\,\pm\,$1.6&74.6$\,\pm\,$2.2&74.4$\,\pm\,$2.5 & 79.8$\,\pm\,$1.0&73.6$\,\pm\,$1.0&72.2$\,\pm\,$1.4 \\
& \textbf{MSA-CNN (small)} & 2024 & \textbf{79.8}$\,\bm{\pm}\,$\textbf{0.9}&\textbf{76.8}$\,\bm{\pm}\,$\textbf{1.1}&\textbf{73.2}$\,\bm{\pm}\,$\textbf{1.1} & \underline{83.9$\,\pm\,$0.4}&\underline{77.3$\,\pm\,$0.5}&\underline{77.8$\,\pm\,$0.5} & 80.8$\,\pm\,$0.7&72.4$\,\pm\,$1.4&72.8$\,\pm\,$1.1 \\
& \textbf{MSA-CNN (large)} & 2024 & \underline{78.6$\,\pm\,$0.4}&\underline{75.3$\,\pm\,$0.5}&\underline{71.6$\,\pm\,$0.6} & \textbf{84.4}$\,\bm{\pm}\,$\textbf{0.3}&\textbf{77.7}$\,\bm{\pm}\,$\textbf{0.4}&\textbf{78.4}$\,\bm{\pm}\,$\textbf{0.4} & \textbf{81.4}$\,\bm{\pm}\,$\textbf{0.3}&73.8$\,\pm\,$0.3&\textbf{73.7}$\,\bm{\pm}\,$\textbf{0.4} \\
\bottomrule
\end{tabular}}
    \label{tab:metrics}
\end{table*}
\begin{table*}[]
    \centering
    \caption{Comparison of model complexity in terms of number of parameters and MFLOPS. The lowest model complexity of all models without engineered features is marked in bold; the second-lowest is underlined}
    \resizebox{0.9\textwidth}{!}{
    \begin{tabular}{lcccrrrrrr}
\toprule
model & year & engineered & framework & \multicolumn{2}{c}{univariate} & \multicolumn{2}{c}{multivariate\:--\:ISRUC-S3} & \multicolumn{2}{c}{multivariate\:--\:Sleep-EDF} \\ \cmidrule(lr){5-6}\cmidrule(lr){7-8}\cmidrule(lr){9-10}
& & \smash{\raisebox{0.5em}{features}} & & \# parameters & MFLOPS & \# parameters & MFLOPS & \# parameters & MFLOPS \\
\midrule
GraphSleepNet \cite{jia2020graphsleepnet}& 2020 & \ding{51} & Keras & --\hspace{2.75pt} & --\hspace{2.75pt} & 2,740&0.03 & 1,552&0.009 \\ \cdashlinelr{1-10}
DeepSleepNet \cite{supratak2017deepsleepnet}& 2017 & \ding{55} & PyTorch &  35,912,074&48.2 & --\hspace{2.75pt} & --\hspace{2.75pt} & --\hspace{2.75pt} & --\hspace{2.75pt} \\
EEGNet \cite{lawhern2018eegnet}& 2018 & \ding{55} & PyTorch &  \underline{12,037}&9.5 & \underline{12,181}&32.8 & \underline{12,085}&17.2 \\
AttnSleep \cite{eldele2021attention}& 2021 & \ding{55} & PyTorch & 522,805&122.5 & --\hspace{2.75pt} & --\hspace{2.75pt} & --\hspace{2.75pt} & --\hspace{2.75pt} \\
MSTGCN \cite{jia2021multi}& 2021 & \ding{55} & Keras & --\hspace{2.75pt} & --\hspace{2.75pt} & 476,292&204.0 & 376,928&81.6 \\
JK-STGCN \cite{ji2022jumping}& 2022 & \ding{55} & Keras & --\hspace{2.75pt} & --\hspace{2.75pt} & 458,085&208.6 & 431,973&83.3 \\
HierCorrPool \cite{wang2023multivariate}& 2023 & \ding{55} & PyTorch &  8,150,000&82.9 & 13,290,000&499.0 & 8,930,000&146.1 \\
FC-STGNN \cite{wang2024fully}& 2024 & \ding{55} & PyTorch &  457,990&46.2 & 3,200,000&475.8 & 1,370,000&186.6 \\
cVAN \cite{yang2024cvan}& 2024 & \ding{55} & Keras & --\hspace{2.75pt} & --\hspace{2.75pt} & 6,834,571&483.4 & 5,699,173&273.4 \\
\textbf{MSA-CNN (small)} & 2024 & \ding{55} & PyTorch &  \textbf{8,517}&\textbf{3.1} & \textbf{10,583}&\textbf{19.8} & \textbf{8,013}&\textbf{9.2} \\
\textbf{MSA-CNN (large)} & 2024 & \ding{55} & PyTorch &  35,301&\underline{8.0} & 43,511&\underline{29.0} & 33,261&\underline{15.3} \\
\bottomrule
\end{tabular}}
    \label{tab:parameters}
\end{table*}

\begin{figure*}[h]
    \centering
    \begin{tikzpicture}
        \sffamily
        \node (image) at (0,0) {\includegraphics[width=\textwidth, trim={0.8cm 0.0cm 0 0},clip]{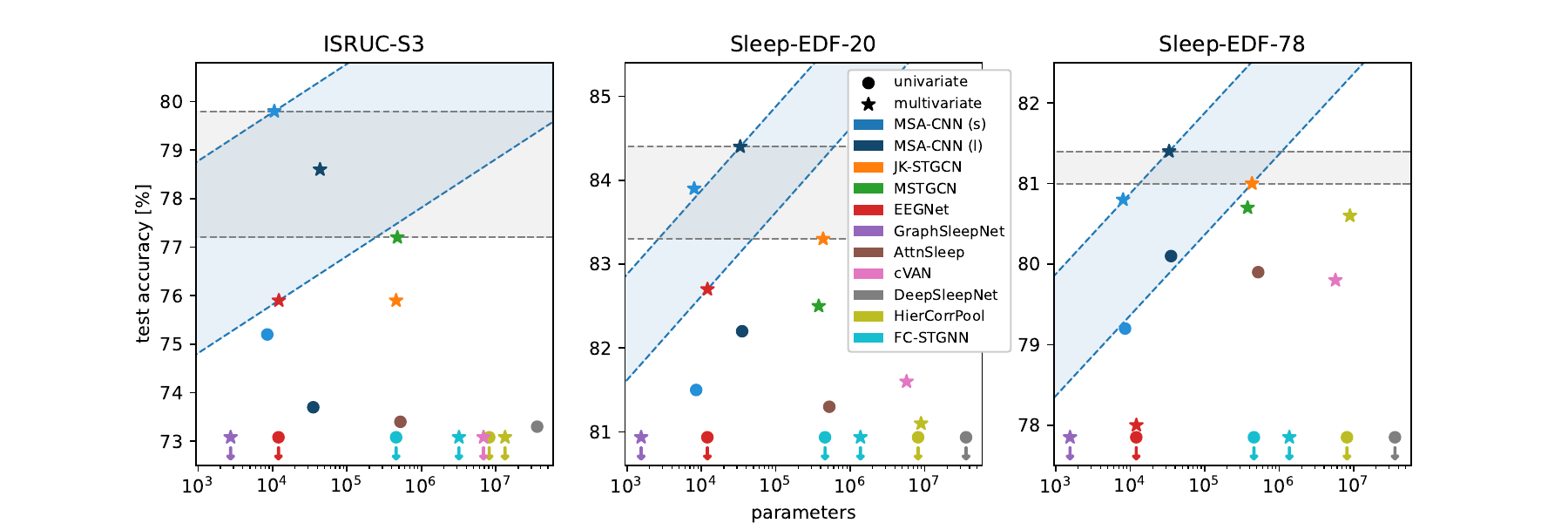}};
        \node (caption) at (-10 + 3.08 - 0.2,2.4+0.12){A}; 
        \node (caption) at (-10+7.95 - 0.2,2.4+0.12){B};
        \node (caption) at (-10+12.85- 0.2,2.4+0.12){C};
    \end{tikzpicture}
    \vspace{-0.3cm}
    \caption{Model performance as a combination of both test accuracy and model complexity on the ISRUC-S3 (A), Sleep-EDF-20 (B) and Sleep-EDF-78 (C) dataset. The two metrics can be combined by defining a balance between them, for example allowing a 10-fold increase in complexity for each percentage point increase in accuracy. This combined performance measure reveals a significant performance gap (blue) across all datasets between our model and the baseline models. Notably, even when excluding model complexity, our MSA-CNN model still outperforms all baseline models, as indicated by the performance gap in grey.}
    \label{fig:balance}
\end{figure*}

\section{Experimental Results}
\label{sec:comparison}

\subsection{Comparison to state-of-the-art (SOTA) models}
Our experimental results are summarised in Table \ref{tab:metrics}. All models are grouped by either univariate or multivariate. 
Our proposed small MSA-CNN model is the best-performing model on the ISRUC-S3 dataset in terms of all tested metrics, irrespective of its configuration as univariate or multivariate. The multivariate version achieved the highest accuracy with 79.8$\pm$0.9\,\%, surpassing the MSTGCN as the best baseline significantly by 2.6 percentage points. On the Sleep-EDF-20 dataset, the multivariate version exceeded every tested baseline.
Conversely, our proposed large MSA-CNN is the best-performing model on the two Sleep-EDF datasets in terms of accuracy and Cohen's kappa, both as a univariate and a multivariate model. 
The multivariate configuration achieves an accuracy of 84.4$\pm$0.3\,\% on the Sleep-EDF-20 dataset, exceeding the best baseline, JK-STGCN, by 1.1 percentage points. The same configuration yields an accuracy of 81.4$\pm$0.3\,\% on the Sleep-EDF-78 dataset, surpassing the best baseline JK-STGCN by 0.4 percentage points. 
While surpassed by the small MSA-CNN, the large, multivariate configuration still outperforms every baseline model on the ISRUC-S3 dataset.
In terms of macro F1 score, the multivariate MSA-CNN is outperformed by the MSTGCN and the JK-STGCN on the Sleep-EDF-78 dataset.
Note that for some of the baselines, such as the AttnSleep model or the graph neural network-based models, the re-evaluated results differ from previously reported results due to the use of our comprehensive validation strategy.

The test accuracy of the large MSA-CNN of 78.6\,\% on the AASM-labelled ISRUC-S3 dataset is about 4 percentage points below the AASM inter-scorer agreement of 82.6\,\% obtained by Rosenberg et al. \cite{rosenberg2013american}.
On the other hand, the test accuracies of 84.4\,\%, and 81.4\,\% on the respective R\&K-labelled Sleep-EDF-20 and Sleep-EDF-78 datasets exceed the inter-scorer agreement of 80.6\,\% reported by Danker-Hopfe et al. \cite{danker2009interrater}. Together, the results suggest that the performance of the large MSA-CNN is comparable to that of sleep experts.

\begin{figure*}[t]
    \centering
    
      \begin{tikzpicture}
      \sffamily
      \node (image) at (0,0) {\includegraphics[width=1\textwidth]{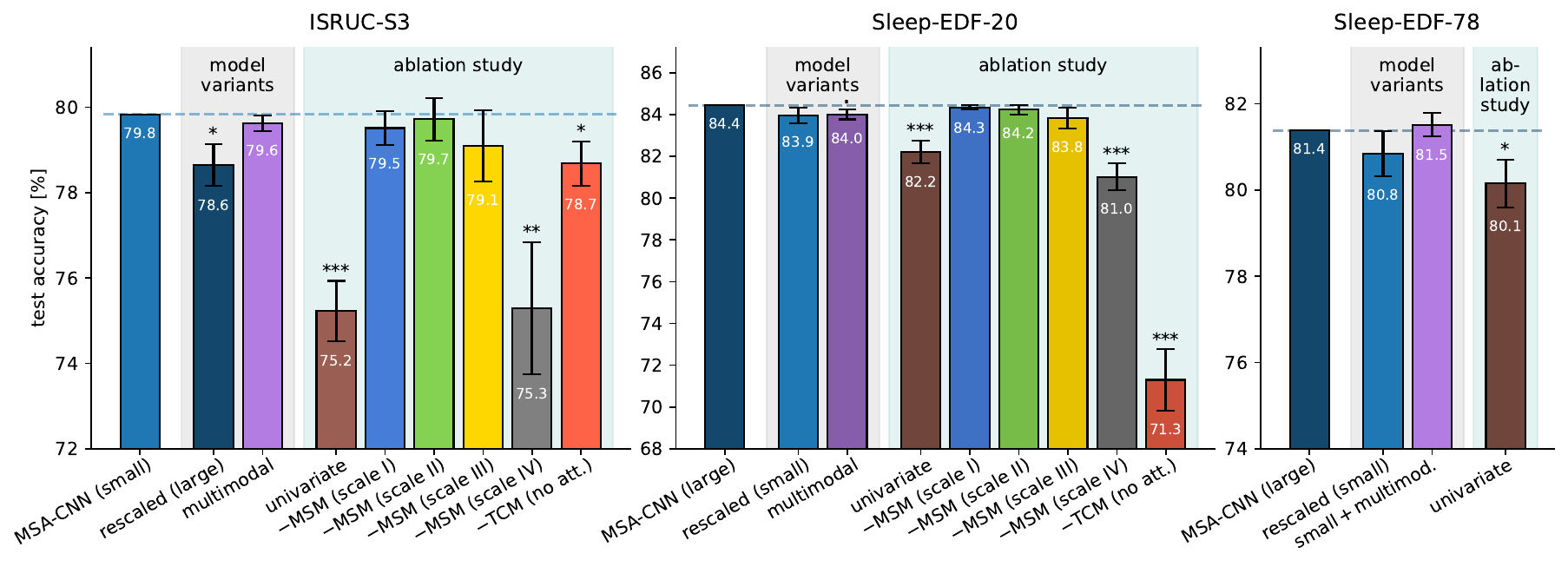}};
      \node (caption) at (-10 + 2.32-0.2,2.8){A}; 
      \node (caption) at (-10 + 8.8-0.2,2.8){B};
      \node (caption) at (-10 + 15.28-0.2,2.8){C};
      \end{tikzpicture}
    \vspace{-0.5cm}
    \caption{Performance of model variants and ablation study on the datasets ISRUC-S3 (A), Sleep-EDF-20 (B), and Sleep-EDF-78 (C) in terms of test accuracy. The proposed MSA-CNN model is configured as small (bright blue) for the ISRUC-S3 dataset and as large (dark blue) for the larger Sleep-EDF datasets, with the complementary model size serving as a model variant. 
    For the ablation study, we changed the model to univariate (brown), replaced the multi-scale convolutions with a single uni-scale convolution (scale colour), or removed attention from the proposed model (red). Light (dark) colours indicate a modification from the small (large) MSA-CNN. Error bars depict the standard error of the mean across folds paired with the proposed model, while significant deviations from the proposed model, established using a paired $t$-test, are indicated above the error bar ({$\cdot$}: $p<0.1$, {$*$}: $p<0.05$, {$**$}: $p<0.01$, {$**$$*$}: $p<0.001$).}
    \label{fig:ablation}
\end{figure*}

\subsection{Model Complexity Analysis}
Table \ref{tab:parameters} presents a comparison of model complexity between our model and SOTA models.
Note that the model complexity depends on the number of input channels, which 
varies between univariate and multivariate configurations and across datasets.

Among all raw input models, the small MSA-CNN shows the lowest model complexity in both the number of parameters and MFLOPS. The large MSA-CNN ranks second in MFLOPS complexity, while its parameter count is the third lowest, behind EEGNet. The model with the fourth lowest number of parameters, the JK-STGCN, has already more than ten times as many parameters as our large MSA-CNN on all datasets. 
The DeepSleepNet exhibits the highest model complexity, with more than 1,000 times as many model parameters as our large MSA-CNN and more than six times as many MFLOPS.

In addition, Figure \ref{fig:balance} shows the model performance as a combination of test accuracy and model parameters. When assigning a 10-fold reduction in complexity a value of one percentage point in accuracy, the combined metric reveals a substantial gap between the multivariate MSA-CNN models and all baselines, irrespective of model size configuration and dataset.

\subsection{Model Variants and Ablation Study}
\label{ssec:ablation}
We explored two variants of our MSA-CNN model and conducted an ablation study for three proposed techniques and modules.
The small MSA-CNN serves as the reference model for the smaller ISRUC-S3 dataset, while the large MSA-CNN serves as the reference model for the Sleep-EDF datasets.
The two model variants are:
\begin{enumerate}
    \item \textbf{rescaled}: large MSA-CNN for the ISRUC-S3 dataset and small MSA-CNN for the Sleep-EDF datasets.
    \item \textbf{multimodal}: filters are learnt separately for each channel (see Subsection \ref{ssec:msm}).
\end{enumerate}
In addition, the three ablation variants are:
\begin{enumerate}
    \setcounter{enumi}{2}
    \item \textbf{univariate}: single-channel variant, as described in Subsection \ref{ssec:modelconfig}.
    \item \textbf{$-$MSM}: MSM replaced with a single-scale layer. The model size is maintained by matching the number of overall filters. All four scales from Table \ref{tab:scales} are tested.
    \item \textbf{$-$TCM}: TCM removed from the model. This variant passes twice as many features to the fully connected layer, as the TCM reduces the feature dimension by half.
\end{enumerate}

\noindent
\begin{table}[]
    \centering
    \caption{Model complexity of model variants and ablation study. Bold font marks the model complexity of the reference model used in the ablation study for the respective dataset}
    \resizebox{1\columnwidth}{!}{%
    \begin{tabular}{llrrrr}
\toprule
& model &  \multicolumn{2}{c}{ISRUC-S3}&\multicolumn{2}{c}{Sleep-EDF} \\ \cmidrule(lr){3-4}\cmidrule(lr){5-6}
& & \# parameters & MFLOPS & \# parameters & MFLOPS\\
\midrule
\multirow{4}*{\rotatebox[origin=c]{90}{variants}}&MSA-CNN (small) & \textbf{10,583} & \textbf{19.8} & 8,013 & \phantom{0}9.2\\
&multimodal (small) & 13,327 & \phantom{0}9.9 & 7,517 & \phantom{0}5.1\\
&MSA-CNN (large) & 43,511 & 29.0     & \textbf{33,261} & \textbf{15.3}\\
&multimodal (large) & 42,599 & 15.1 & 29,709 & 10.7\\
\cdashlinelr{1-6}
\multirow{6}*{\rotatebox[origin=c]{90}{ablation}}& univariate & 8,517 & \phantom{0}3.1 & 35,301 & \phantom{0}8.0\\
&$-$MSM (scale I) & 10,583 & 36.2 & 33,261 & 24.1\\
&$-$MSM (scale II) & 10,583 & 20.6 & 33,261 & 17.2\\
&$-$MSM (scale III) & 10,583 & 12.8 & 33,261 & 13.8\\
&$-$MSM (scale IV) & 10,583 & \phantom{0}8.9 & 33,261 & 12.0\\
&$-$TCM (no att.) & 7,911 & 19.0 & 14,253 & 10.9\\
\bottomrule
\end{tabular}}
    \label{tab:parameters_ablation}
\end{table}

We conducted the testing of the model variants and the ablation study on all three datasets.
However, we excluded the Sleep-EDF-78 dataset from the ablation study of the two modules (4-5) due to its similarity with the Sleep-EDF-20 dataset and computational constraints as a result of its extensive size.
The model variants and ablated models were repeated 10 times (3 times for the Sleep-EDF-78 dataset). 
To compare the performance of the modified models to the reference model, we averaged the test accuracy for each fold across all repetitions and analyzed the resulting paired observations using a linear model with the reference configuration as baseline. Error bars correspond to the standard errors of the estimated performance differences.

\begin{figure}
    \centering
    \resizebox{\columnwidth}{!}{
    \begin{tikzpicture}
        \sffamily
        \node (image) at (0,0) {\includegraphics[width=\columnwidth, trim={0.255cm, 0cm, 0.28cm, 0.22cm}, clip]{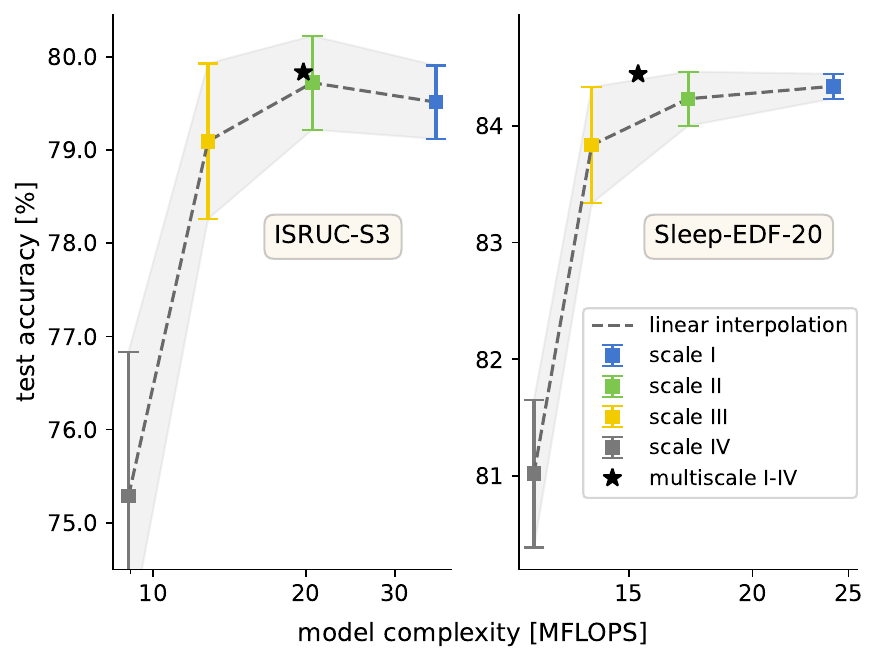}};
        \node (caption) at (-5 + 1.785 -0.21,3.355){A}; 
        \node (caption) at (-5+5.8 -0.21,3.355){B};
    \end{tikzpicture}
    }
    \vspace{-0.37cm}
    \caption[Uni-scale complexity analysis MSA-CNN]{Model test accuracy relative to computational complexity for datasets ISRUC-S3 and Sleep-EDF-20. 
    Error bars depict the standard error across repetition-averaged folds. The multi-scale model demonstrates superior performance when jointly considering test accuracy and computational complexity.}
    \label{fig:uniscale}
\end{figure}

The results for the model variants and the ablation study are presented in Figure \ref{fig:ablation}. In addition, Table \ref{tab:parameters_ablation} shows the number of model parameters for each modified model. Regarding model variants, the figure shows that the rescaled large MSA-CNN performs significantly worse than the reference model on the ISRUC-S3 dataset. On the Sleep-EDF datasets, the small MSA-CNN is slightly lower than the reference model. On the other hand, the table shows that the model complexity of the small MSA-CNN is substantially lower than that of the large MSA-CNN. 
The multimodal configuration is slightly lower on the ISRUC-S3 and the Sleep-EDF-20 datasets. Notably, the small multimodal model shows a marginal, albeit not significant, improvement over the reference model, despite having just 7,517 model parameters\thinspace---\thinspace less than one-fourth the number of parameters of the reference model. Overall, the model variants represent promising alternatives to the proposed reference model and should be explored in more detail in future work.

The first \textbf{ablation study} demonstrates that the multivariate approach significantly improves model performance, irrespective of the dataset. Similarly, the multi-scale ablation studies show that the incorporation of the MSM outperforms all uni-scale models.
While the performance improvement of the multi-scale model is only marginal compared to uni-scale models I and II, 
the reduction in MFLOPS model complexity (Table \ref{tab:parameters_ablation}) should also be taken into account.
Figure \ref{fig:uniscale} shows test accuracy as a function of model complexity. Across both the ISRUC-S3 and Sleep-EDF-20 datasets, the multi-scale configuration achieves higher performance than the interpolated uniscale configuration at comparable model complexity.
This effect is more pronounced on the larger of the two datasets, Sleep-EDF-20, where the performance of the multi-scale model lies above the interpolated standard error region.
Lastly, the ablation study for the TCM shows that the TCM improves model performance significantly, proving that the TCM is an integral part of the model.

Although limited to the two smaller datasets, the last two ablation studies show that the TCM contributes significantly to model performance, while the MSM shows modest yet consistent improvements. Taken together, these results support our decision to omit a full ablation on the Sleep-EDF-78 dataset, as the expected additional insight would likely be limited relative to the substantial computational cost.

\begin{figure}
    \centering
    \begin{tikzpicture}
        \sffamily
        \node (image) at (0,0) {\includegraphics[width=0.99\columnwidth, trim={0.31cm, 0cm, 1.15cm, 0.80cm}, clip]{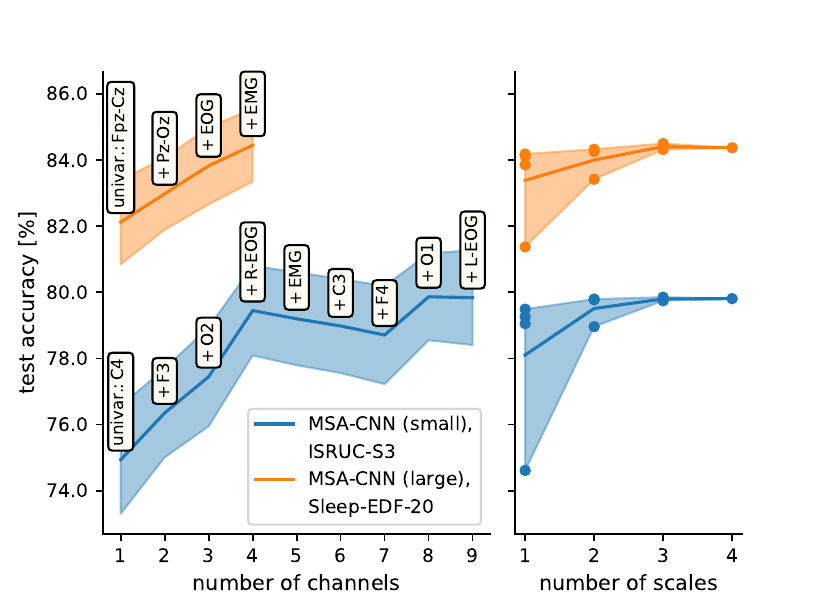}};
        \node (caption) at (-5 + 1.805 -0.21,3.){A}; 
        \node (caption) at (-5+6.425 -0.21,3.){B};
    \end{tikzpicture}
    \vspace{-0.25cm}
    \caption{Test accuracy parameter sensitivity of MSA-CNN small (large) on dataset ISRUC-S3 (Sleep-EDF-20). (A) Mean test accuracy relative to number of channels. The shaded area shows the standard error of the mean across repetition-averaged folds. Starting from a univariate configuration, channels are successively added in a predetermined order.
    (B) Test accuracy relative to the number of contiguous scales in the multi-scale module. Each measurement (circle) depicts a different configuration of contiguous scales. The solid lines show the mean test accuracy, while the shaded area shows the minimum and maximum for each number of scales.}
    \label{fig:channels_scales}
\end{figure}

\subsection{Parameter Sensitivity Analysis}
\label{ssec:parameter_sensitivity}
\color{black}
This section assesses the robustness of the proposed modules and techniques through parameter sensitivity analyses.
Specifically, we analyse the effect of the number of channels as well as of the number of scales in the MSM on model performance. 
We set the small MSA-CNN as the reference model for the ISRUC-S3 dataset, and the large MSA-CNN for the Sleep-EDF-20 dataset. Similar to the ablation study, we excluded the Sleep-EDF-78 dataset due to its redundancy with the Sleep-EDF-20 dataset and its larger size.

For the multivariate approach, we conducted a parameter sensitivity analysis by varying the number of input channels from a single channel (univariate) to the full set of channels. The set of channels is determined by successively adding channels to the single-channel configuration. We aim to prioritise channels that provide the most relevant information first, such as channels from not yet included brain regions or contralateral hemispheres.
Note that we employed the multivariate model configuration even for univariate inputs to ensure a direct comparison of the number of channels. Each configuration is repeated 10 times and each fold is averaged across the repetitions. Furthermore, mean and standard error are computed across the averaged folds. 
The results are presented in Figure \ref{fig:channels_scales} (A). 
For the ISRUC-S3 dataset, the model performance initially increases with the number of channels up to four channels, but levels off after roughly four channels with the addition of the EMG signal. Note that the remaining channels are contralateral counterparts of already included channels.
For the Sleep-EDF-20 dataset, the model performance increases with the number of channels up to the full set of four channels. 
Note that all channels of this dataset are medially located.

\begin{figure*}[t]
    \centering
    \includegraphics[width=0.99\textwidth, trim={2.5cm, 1cm, 3cm, 1cm}, clip]{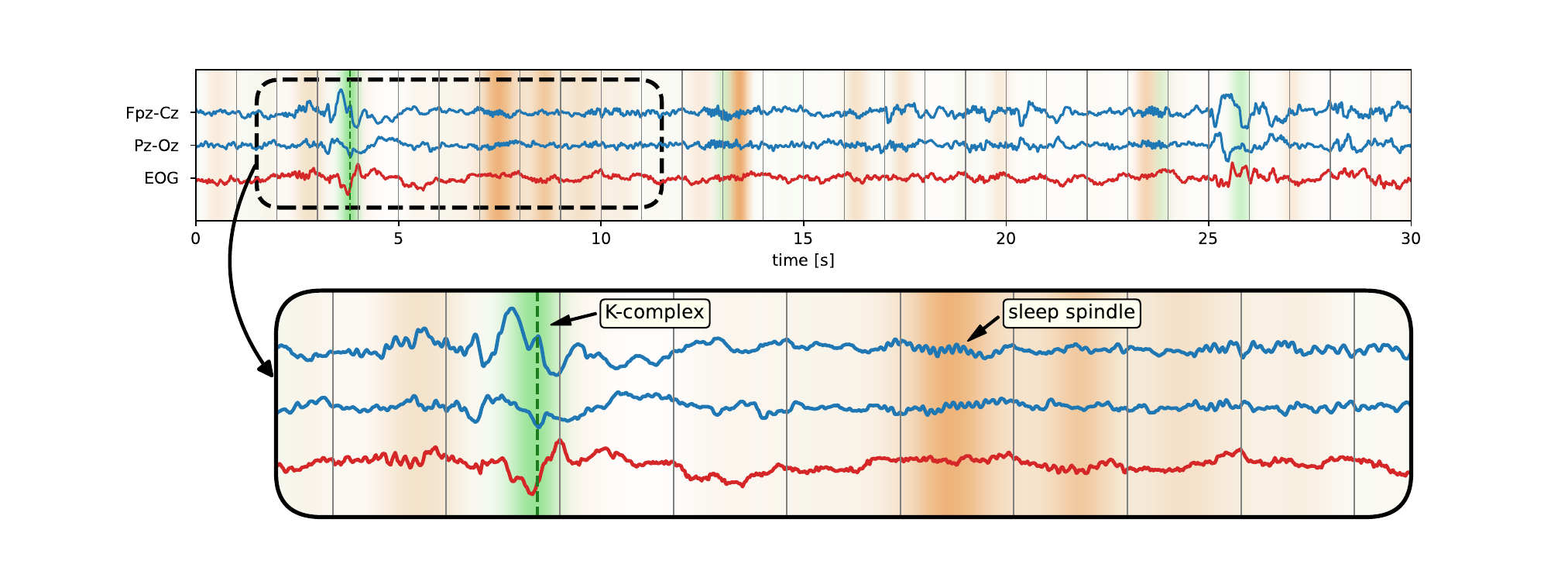}%
    \caption{Demonstration of the attention mechanism in our experiment. The Sleep-EDF-20 sample, labelled as sleep stage N2, is taken from one of the participants assigned to the test set. 
    Only the EEG- and EOG-based channels of the multivariate signal are shown.
    We trained our large MSA-CNN and derived the attention matrix weights for this sample. Areas shaded in green (incoming attention) show the extent to which this area is attended by all other matrices on average, with a maximum at roughly 3.8 seconds (marked in green). Orange (outgoing attention) depicts the extent to which this area attends to the maximum (green dashed line). The zoom-in inset magnifies the sample by a factor of three and shows a K-complex attended by a sleep spindle, which is a common occurrence in sleep stage N2. The model correctly classified the sample as sleep stage N2 with a prediction probability above 99.9\,\%.}
    \label{fig:attention_visualisation}
\end{figure*}

To perform a parameter sensitivity analysis on the MSM, we systematically reduce the number of scales. Specifically, one of the two outermost scales is successively removed, ensuring that the remaining scales are contiguous. At the same time, the number of filters per scale is increased to maintain the total number of multi-scale filters, which also ensures no change to the overall number of model parameters. This approach yields two configurations with three scales, three configurations with two scales, and four uni-scale configurations with one scale, the latter of which are identical to the base models used in the ablation study.
Each scale configuration was repeated 10 times on the ISRUC-S3 dataset and 3 times on the Sleep-EDF-20 dataset. In the case of no model modification (four scales), the results are drawn from our main study (Section \ref{sec:comparison}), which includes 10 repetitions. Lastly, the average configuration accuracy across folds and repetitions is computed. 
Figure \ref{fig:channels_scales} (B) shows the results of the sensitivity on the number of MSM scales. On average, the performance increases with the number of scales for both datasets. While the best-performing configuration for a given number of scales increases only slightly, the gap between the best- and worst-performing configurations with a given number of scales narrows significantly as the number of scales increases.

\subsection{Visualisation of Attention}
\label{ssec:visualattention}
\color{black}

In Figure \ref{fig:attention_visualisation}, we visualise the attention mechanism of the large MSA-CNN for an N2 sleep stage sample from the Sleep-EDF-20 dataset, using the technique proposed in Subsection \ref{ssec:TCM}.
The figure shows the EEG and EOG signals over the full length of the sample along with the average incoming attention and the outgoing attention relative to the incoming attention maximum. The zoom-in inset focusses on a 10-second window surrounding the incoming attention maximum. It captures how a K-complex is attended by a sleep spindle, visually explaining the function of the attention mechanism in the network. 

\FloatBarrier

\section{Summary \& Conclusion}
\label{sec:summary_conclusion}
In this study, we introduced a lightweight sleep stage classification architecture called MSA-CNN, which 
extends previous model complexity limitation efforts 
to multivariate models.
%
%
Using a comprehensive and transparent validation strategy, the
results demonstrate that the MSA-CNN outperforms all tested SOTA models across three benchmark datasets, reaching a performance comparable to that of human experts. 
At the same time, the MSA-CNN has a lower model complexity than seven of the nine SOTA models.
We hypothesise that the reduction in model complexity may act as a regulariser and simplify training, thereby contributing to the observed performance gains. 
Furthermore, the lower parameter count may strengthen the interpretability of our model and make it more suitable for deployment in practical applications based on mobile or wearable devices. 
These observations underscore the potential of controlling model complexity for sleep stage classification.

To validate the components of our model, we conducted an ablation study and parameter
sensitivity analyses.
These show that incorporating additional channels can significantly improve performance, although this must be balanced against the practical benefits of using fewer measurement channels.
The ablation study further demonstrates that capturing contextual information with attention is an integral component of the MSA-CNN. 

We provided explainability for our model by visualising incoming and outgoing attention, allowing us to uncover the relationships between waveform patterns used by the model for decision making.
This explainability component is complemented by parallel work on explainability of early-stage spectral processing in our model \cite{goerttler2025retrieving}.
In future work, we aim to conduct a systematic evaluation of our introduced attention visualisation method, with the goal of establishing clearer links between observed attention maps and established physiological patterns.

Our current model relies on supervised learning and is therefore affected by inter-scorer variability. 
Future work will focus on developing unsupervised learning models to eliminate reliance on human sleep scoring. 
In addition, a limitation of our study is the absence of medical conditions in the training data, such as sleep apnea. Since sleep staging is especially relevant in clinical contexts, future research should validate the model for these conditions.

\printcredits

\section*{Declaration of competing interest}
The authors declare that they have no known competing financial interests or personal relationships that could have appeared to influence the work reported in this paper.

\section*{Acknowledgments}
Stephan Goerttler was supported by A*STAR ARAP Scholarship.
Fei He was supported by EPSRC grant [EP/\allowbreak X020193/\allowbreak1].

\section*{Data availability}
The ISRUC-S3 dataset that supports the findings of this study is openly available at \url{https://sleeptight.isr.uc.pt/} (Subgroup\_3).  
The Sleep-EDF datasets are available at PhysioNet \url{https://www.physionet.org/content/sleep-edfx/1.0.0/} (Sleep Cassette).

\bibliographystyle{cas-model2-names}

\bibliography{cas-refs}

\end{document}